\documentclass[sigconf]{acmart}

\usepackage{geometry}
\usepackage{graphicx}
\usepackage{float}
\usepackage{caption}
\usepackage{subfig}
\usepackage{bm}
\usepackage{booktabs}
\usepackage{hyperref}
\AtBeginDocument{%
	\providecommand\BibTeX{{%
			\normalfont B\kern-0.5em{\scshape i\kern-0.25em b}\kern-0.8em\TeX}}}

\copyrightyear{2022} 
\acmYear{2022} 
\setcopyright{rightsretained} 

\acmConference[KDD '22]{Proceedings of the 28th ACM SIGKDD Conference on Knowledge Discovery and Data Mining}{August 14--18, 2022}{Washington, DC, USA}
\acmBooktitle{Proceedings of the 28th ACM SIGKDD Conference on Knowledge Discovery and Data Mining (KDD '22), August 14--18, 2022, Washington, DC, USA}
\acmISBN{978-1-4503-9385-0/22/08}
\acmDOI{10.1145/3534678.3539129}

\usepackage{etoolbox}
\makeatletter
\patchcmd{\maketitle}{\@copyrightpermission}{
   \begin{minipage}{0.3\columnwidth}
     \href{https://creativecommons.org/licenses/by/4.0/}{\includegraphics[width=0.90\textwidth]{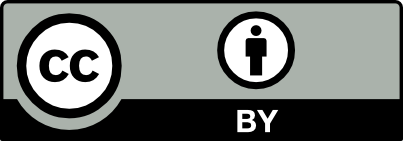}}
   \end{minipage}\hfill
   \begin{minipage}{0.7\columnwidth}
     \href{https://creativecommons.org/licenses/by/4.0/}{This work is licensed under a Creative Commons Attribution International 4.0 License.}
   \end{minipage}
  
   \vspace{5pt}
}{}{}

\makeatother

\begin{document}
	
	\title{Company-as-Tribe: Company Financial Risk Assessment on Tribe-Style Graph with Hierarchical Graph Neural Networks}
	
	\author{Wendong Bi}
	\affiliation{%
		\institution{Institute of Computing Technology, University of Chinese Academy of Sciences}
		\city{Beijing}
		\country{China}}
	\email{biwendong20g@ict.ac.cn}
	
	\author{Bingbing Xu}
	\authornote{Corresponding authors}
	\affiliation{%
		\institution{Institute of Computing Technology, Chinese Academy of Sciences}
		\city{Beijing}
		\country{China}}
	\email{xubingbing@ict.ac.cn}

	\author{Xiaoqian Sun}
	\authornotemark[1]
	\affiliation{%
		\institution{Institute of Computing Technology, Chinese Academy of Sciences}
		\city{Beijing}
		\country{China}}
	\email{sunxiaoqian@ict.ac.cn}
	
	\author{Zidong Wang}
	\affiliation{%
		\institution{Institute of Computing Technology, Chinese Academy of Sciences}
		\city{Beijing}
		\country{China}}
	\email{wangzidong@ict.ac.cn}
	
	
	\author{Huawei Shen}
	\affiliation{%
		\institution{Institute of Computing Technology, Chinese Academy of Sciences}
		\city{Beijing}
		\country{China}}
	\email{shenhuawei@ict.ac.cn}
	
	\author{Xueqi Cheng}
	\authornotemark[1]
	\affiliation{%
		\institution{Institute of Computing Technology, Chinese Academy of Sciences}
		\city{Beijing}
		\country{China}}
	\email{cxq@ict.ac.cn}
	
	\renewcommand{\shortauthors}{Wendong Bi et al.}
	
	\begin{abstract}
		
		Company financial risk is ubiquitous and early risk assessment for listed companies can avoid considerable losses. 
		Traditional methods mainly focus on the financial statements of companies and lack the complex relationships among them. However, the financial statements are often biased and lagged, making it difficult to identify risks accurately and timely.
		To address the challenges, we redefine the problem as \textbf{company financial risk assessment on tribe-style graph} by taking each listed company and its shareholders as a tribe and leveraging financial news to build inter-tribe connections. Such tribe-style graphs present different patterns to distinguish risky companies from normal ones.
		However, most nodes in the tribe-style graph lack attributes, making it difficult to directly adopt existing graph learning methods (e.g., Graph Neural Networks(GNNs)). 
		In this paper, we propose a novel \textbf{H}ierarchical \textbf{G}raph \textbf{N}eural \textbf{N}etwork (TH-GNN) for \textbf{T}ribe-style graphs via two levels, with the first level to encode the structure pattern of the tribes with contrastive learning, and the second level to diffuse information based on the inter-tribe relations, achieving effective and efficient risk assessment. 
		Extensive experiments on the real-world company dataset show that our method achieves significant improvements on financial risk assessment over previous competing methods. Also, the extensive ablation studies and visualization comprehensively show the effectiveness of our method.

	\end{abstract}

	

	\maketitle

	\section{Introduction}
	
	Company financial risk is ubiquitous in the real-world financial market. Early assessment of risks for the listed company can provide decision support for company managers and investment institutions, thereby avoiding considerable losses.
	
\begin{figure*}[h]
		\centering
		\includegraphics[width=0.86\linewidth, height=8.5cm]{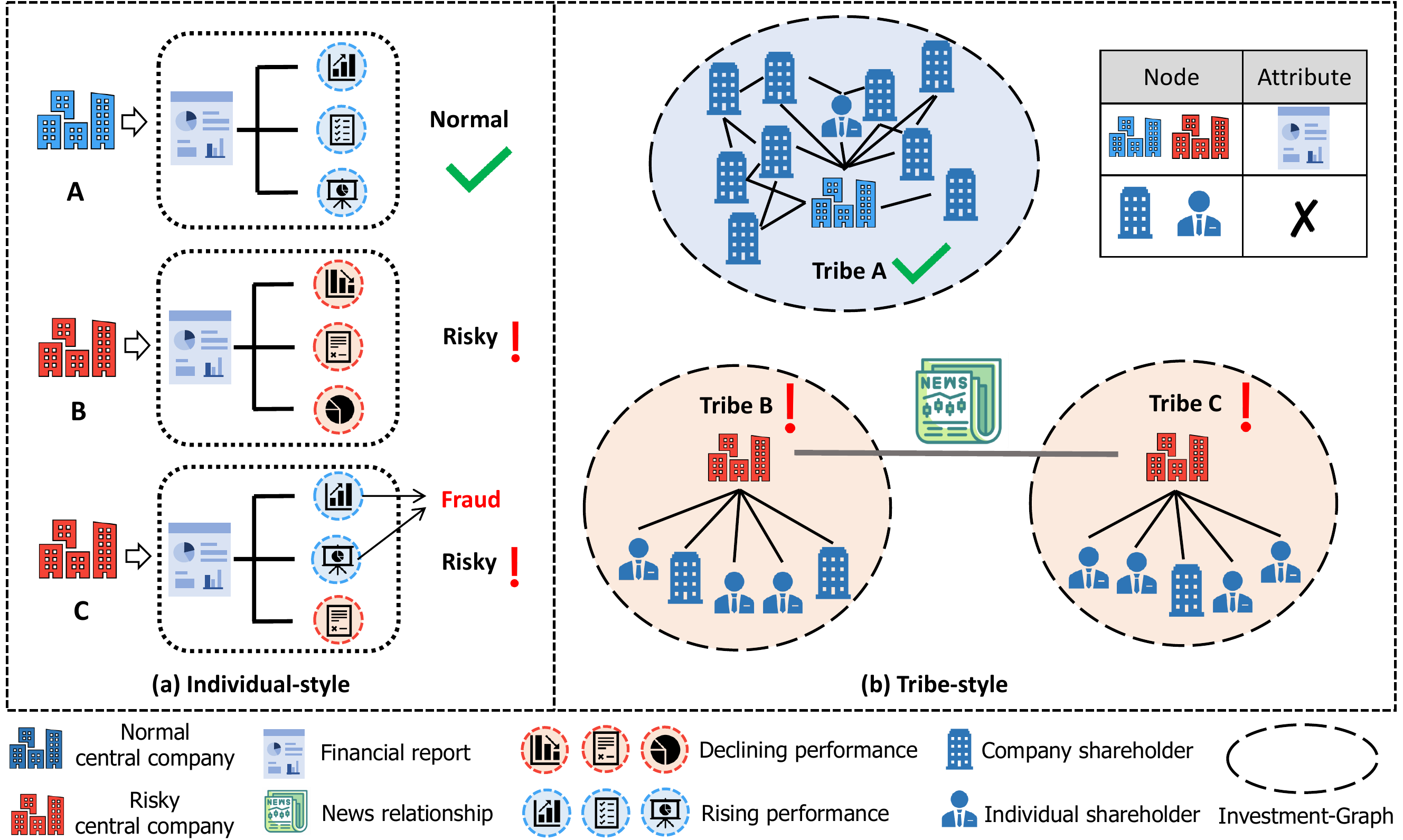}
		\caption{Company financial risk assessment on individual-style vs. tribe-style graph. Each company with its investment-graph in a dotted oval box can be seen as a tribe, and they are further connected by the global relationship (e.g., news relationship).}
		\label{fig:data}
		\Description{Data structure.}
		\vspace{-0.3cm}
	\end{figure*}

	Traditional methods \cite{mai2019deep, chen2020ensemble}, such as financial probability methods, decision tree methods, and Deep Neural Networks (DNNs), treat each company individually and solely leverage the financial statements to assess risk. However, financial statements are often biased and lagged. As Fig.~\ref{fig:data} (a) shows, most companies beautify their published financial data, and some companies even commit financial fraud. Besides, these traditional methods ignore the interactions among companies, which is critical because risks can passed between companies. The above limitations make the traditional methods difficult to identify risks accurately and early.
	
	
	To effectively assess company financial risks, we found there exist two other types of valuable information: 1) the investment-graph of listed companies, e.g., CATL$\footnote{CATL (Contemporary Amperex Technology Co., Limited) is a typical listed company in China, which is a global leader of new energy innovative technologies and committed to providing premier solutions and services for new energy applications worldwide.}$ has more than 200 investors (companies or individuals), which form an investment-graph. As Fig.~\ref{fig:data} (b) shows, risky and normal companies often have different investment patterns; 2) The news-graph among listed companies, i.e., there exists an edge between two companies if they concurrent in at least one piece of news. As Fig.~\ref{fig:data} (b) shows, two listed companies connected usually have strong correlations, and risks can spread over this graph. Superior to other information, financial news is objective and can timely reflect risks. Extensive statistical analyses are provided in Sec.~\ref{section:data_analysis} to demonstrate the benefit of these data for distinguishing risky companies from the normal ones.
	
	Based on the above findings, we redefine the problem as \textbf{company financial risk assessment on tribe-style graphs}. As illustrated in Fig.~\ref{fig:data} (b), we take the investment-graph consisting of a central listed company and its shareholders as a tribe, and leverage the news-graph to construct inter-tribe edges, the financial statements of listed companies are regarded as initial attributes of tribes. 
	However, it is challenging to directly adopt existing graph methods (e.g., Graph Neural Networks (GNNs)) to such tribe-style graphs due to the following serious issues: 1) only the listed companies in a tribe have attributes, and other individuals or companies have no disclosure obligation and therefore do not have attributes, making it difficult to conduct message passing in GNNs; 2) The whole tribe-style graph including both intra-tribe and inter-tribe relationships is large-scale and contains millions of edges, which makes the GNNs inefficient, and traditional node sampling techniques \cite{GraphSAGE, chen2017stochastic, huang2018adaptive}  cause the loss of  information.

	In this paper, we propose a novel \textbf{H}ierarchical \textbf{G}raph \textbf{N}eural \textbf{N}etwork for the financial risk assessment on the \textbf{t}ribe-style graph, namely TH-GNN. Specifically, for the first challenge that the individuals and non-central companies in a tribe have no attributes, we find the structure patterns can reflect the company's risks. Therefore, we design a tribe structure encoder (TSE) based on contrastive learning that learns structural patterns for each tribe (including the scale of the tribe and its investment structure, etc.) without relying on node attributes. For the second challenge, although the whole graph is huge, Fig.~\ref{fig:data} (b) shows an important property of the tribe-style graph that the intra-tribe connections (investment-graph) are dense while the inter-tribal connections (news-graph) are sparse. Inspired by this property, TH-GNN encodes the tribe-style graph through a hierarchical manner, with the first level to encode tribes defined by the investment graphs via the tribe structure encoder and the second level to diffuse the information among tribes based on the news-graph and learn the global representations. Unlike the traditional GNNs that diffuse information over edges on the whole graph, TH-GNN converts a tribe-style graph into parallelly computable local graphs and a smaller global graph.
	Extensive experiments on the real-world dataset for company financial risk assessment show that our approach achieves significant improvement over previous competing methods. Meanwhile, the ablation studies and visualizations also comprehensively show the effectiveness of our method.
	
	The main contributions of this work are summarized as follows:
	
	\begin{enumerate}
		\item We redefine the previous individual risk assessment problem as \textbf{company financial risk assessment on tribe-style graph} and further design a tribe-style graph consisting of financial statements, investment-graphs, and news-graphs rather than solely utilizing financial statements.
		\item We propose a novel Hierarchical Graph Neural Network named TH-GNN to model the tribe-style graph. To the best of our knowledge, this is the first graph representation learning method for company financial risk assessment on the tribe-style graph.
		\item We conduct extensive experiments on a real-world company graph dataset with 0.88 million nodes and 1.31 million edges. The results demonstrate the superiority of the proposed model over state-of-the-art methods. The code is avaliable \footnote{Our source code is available at \href{https://github.com/wendongbi/TH-GNN}{https://github.com/wendongbi/TH-GNN}}.
	\end{enumerate}

	\section{Preliminary}
	\label{sec:priliminary}
	In this section, we present the detailed statistical analysis of tribe-style graph and the formalized definition of company financial risk assessment on tribe-style graphs.
	\subsection{Data Analysis}
	\label{section:data_analysis}
	As illustrated in Fig.~\ref{fig:data} (b), the tribe-style graph consists of investment-graph (tribe) and news-graph, where the investment-graph presents the intra-tribe connections and the news-graph presents the inter-tribe connections. We then analyze the investment-graphs and news-graph comprehensively to verify their benefits for financial risk assessment.
	\subsubsection{investment-graph analysis}
	\label{section:local_graph_analysis}
	An investment-graph usually consists of one central listed company and others (companies or individuals) which have investment relationships with the central company. Only the listed company publishes its financial statements as attributes, and the other nodes do not have attributes. Therefore we mainly focus on the structure patterns of tribes. Then we first give a case study of investment-graphs and then present the statistical analysis on all listed companies' investment-graphs.
	\begin{figure}[h]
		\begin{minipage}[t]{0.5\linewidth}
			\centering
			\subfloat[A risky company example.]{\includegraphics[width=\linewidth, height=2.8cm]{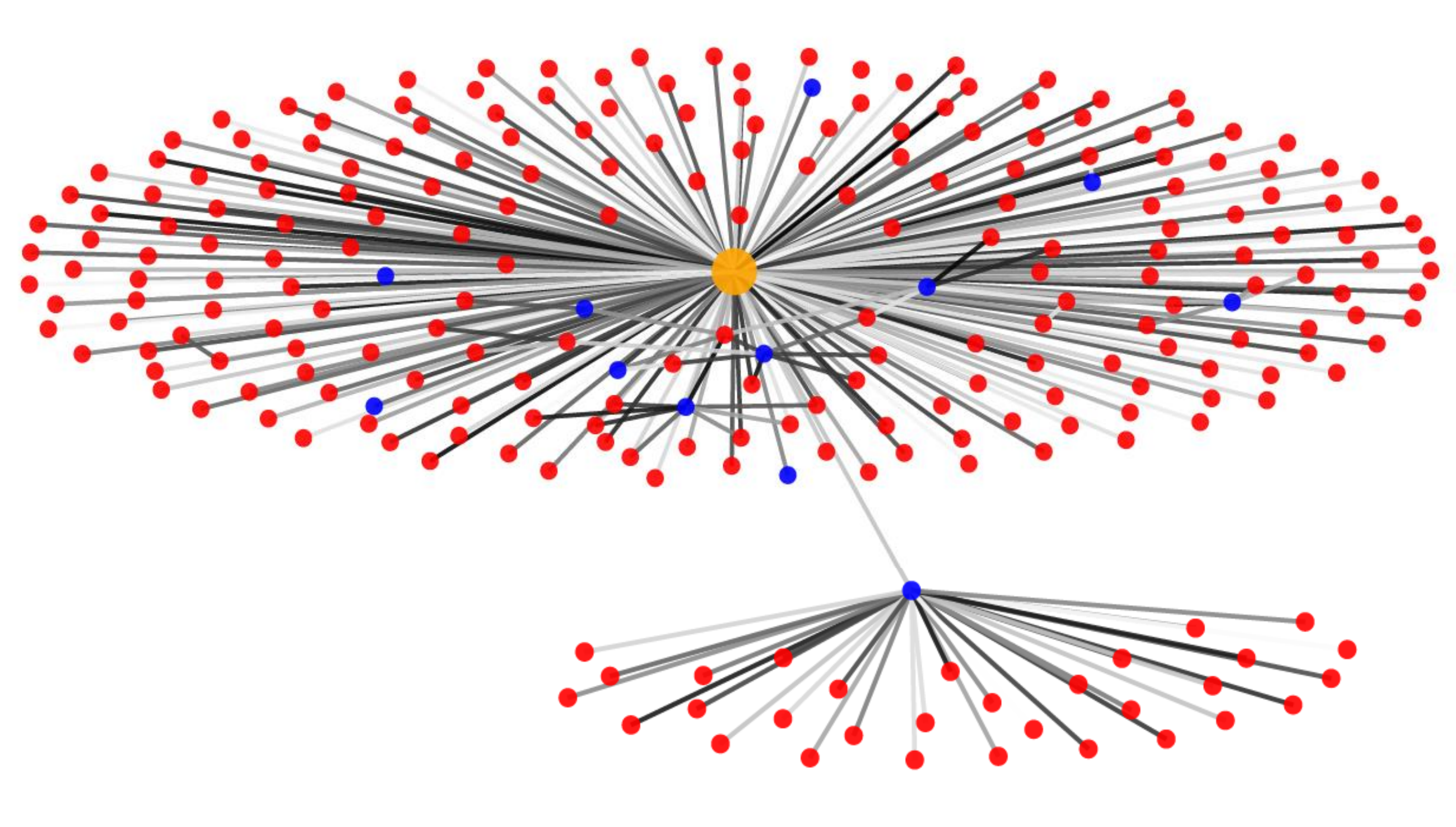}}
		\end{minipage}%
		\begin{minipage}[t]{0.5\linewidth}
			\centering
			\subfloat[A normal company example]{\includegraphics[width=\linewidth, height=2.8cm]{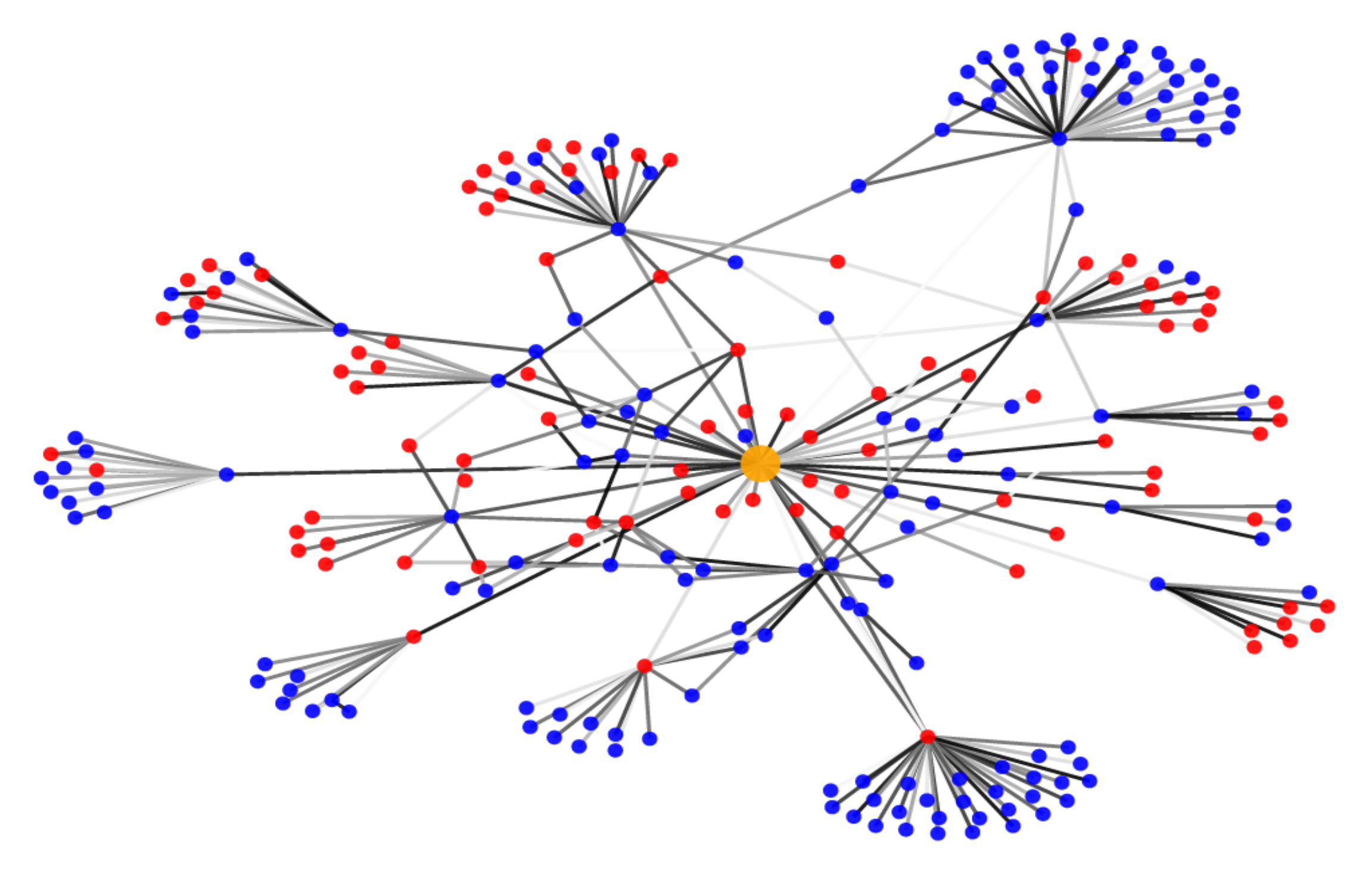}}
		\end{minipage}
		\caption{Examples of investment-graphs\protect\footnotemark[3]. (The bigger orange point denotes listed company, blue points denote unlisted companies and red points denote individuals. And the edges in the graphs represent investment relationships.)}
		
		\label{fig:subgraph_examples}
	\end{figure}
	\footnotetext[3]{These two examples are from two real-world companies named LongYuan Construction and ShangHai Dragon Corporation repectively. }
	As illustrated in Fig.~\ref{fig:subgraph_examples}, the investment-graphs for risky company and normal company show different patterns. The investment-graph of the risky company shown in Fig.~\ref{fig:subgraph_examples} (a) is more similar to a star-like graph, where the single listed company can be viewed as the central node, and its neighboring nodes (investors) consist of more individuals or companies that tend to be disconnected from each other. Different from the risky company, the neighboring nodes of the normal company in Fig.~\ref{fig:subgraph_examples} (b) are often popular companies which have more neighbors (investors), and there exist dense connections among these neighbors. Such a pattern with more reliable investors tends to be more stable, and thus the central listed company is less likely to have financial risks. This motivates us to leverage the investment pattern of listed company to identify financial risks.
	
	To further verify this finding, we conduct centrality analysis for the investment-graphs of all listed companies. Five typical metrics are used \cite{bonacich1987power, newman2018networks}, including degree centrality, eigenvector centrality, clustering centrality, number of bridge, and the central node degree. We calculate the above metrics on each investment-graph and then take the average of risky and normal companies respectively to reflect the differences of centrality between the two classes.
	
	\begin{table}[h]
		\centering
		\caption{Statistical centrality analysis of the investment-graphs}
		\label{tab:centrality}
		\begin{tabular}{ccc}
			\toprule
			Statistical metric & risky company & normal company\\
			\midrule
			Degree centrality & 0.264 & 0.224 \\
			Eigenvector centrality & 0.4161 & 0.3604 \\
			Clustering coefficient & 0.2102 & 0.1907 \\
			Number of bridge (avg) & 125.4 & 112.3 \\
			Central node degree (avg) & 58.43 & 46.71 \\
			\bottomrule
		\end{tabular}
	\end{table}
	
	The results of the centrality analysis are presented in Table~\ref{tab:centrality}, which demonstrate that the investment-graphs of risky companies usually have larger graph centrality compared with that of normal companies.	And this pattern motivates us to take the structure encoding of investment-graph into consideration for financial  risk assessment. We consider the structural pattern of a investment-graph as a tribe  individually, where the nodes within a tribe are centered on the centrally-located listed company. We further explain the benefits of tribe-style graphs in Sec.~\ref{sec:benefit_tribes}.

	
	
	\subsubsection{news-graph analysis}
	\label{section:public_graph_analysis}
	Besides the investment relationship in tribes, we also use financial news to model the interactions between different tribes. The financial news has evident timeliness and objective authenticity, reporting company financial risks promptly. Specifically, different companies may appear in the same news, reflecting strong correlation among them, e.g., news reported that Company A and Company B jointly invested in a failed project, reflecting that they may have potential financial risks at the same time (news used for graph construction in this paper all describes similarity among companies). Then for companies co-existed in one piece of news, we connect them to construct news-graphs, indicating the risky associations among them. 
	
\begin{figure}[h]
		\centering
		\includegraphics[width=0.9\linewidth]{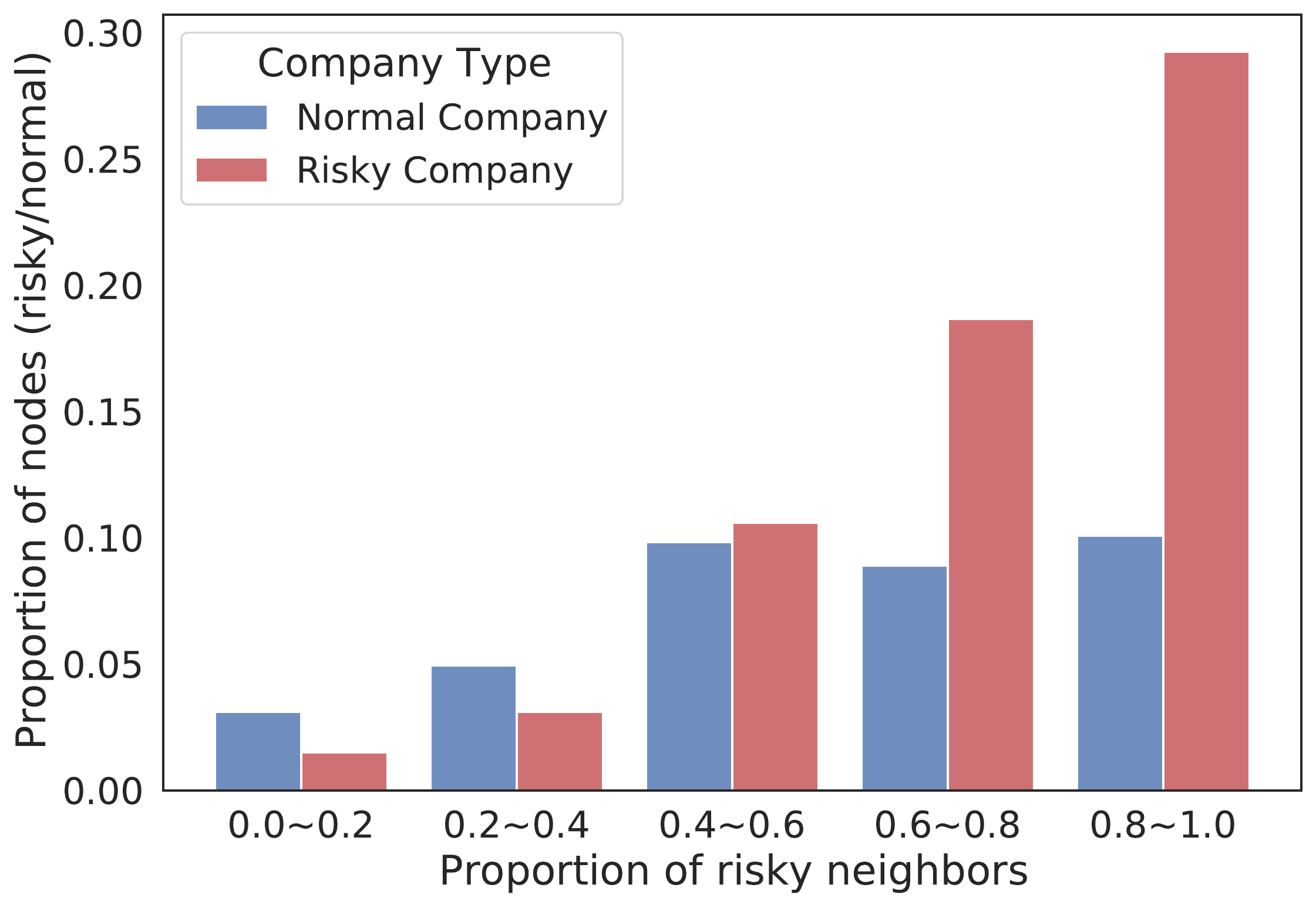}
		\caption{Analysis on proportion of risky neighbors.The x-axis is the proportion of risky companies in all neighbors of each node, and the y-axis is the corresponding proportion of nodes in a certain node type (risky/normal) . Note that we ignore the 0-degree nodes when calculating these results.}
		\label{fig:global}
		\Description{Stack distribution}
	\end{figure}

	As illustrated in Fig.~\ref{fig:global}, we also conduct statistical analysis to validate the benefits of the news-graph. We calculate the proportion of risky companies in neighbors of each node for risky companies and normal companies respectively. For example, the rightmost column in Fig.~\ref{fig:global} indicates there are nearly 30\% of risky companies (the red bar) with more than 80\% of neighbors at-risk, while no more than 10\% of normal companies (the blue bar) with more than 80\% of neighbors at-risk.
	The results show that the probability of risky companies with high-proportion risky neighbors is much larger than that of normal companies, that is, the risky nodes in the news-graph have higher-proportional-risk neighbors. This finding also suggests that news-graph can benefit the company financial risk assessment. As a result, we use the news-graph to construct inter-tribe edges, modeling the relations among listed companies.

	\begin{figure*}[h]
		\centering
		\includegraphics[width=0.88\linewidth, height=8.2cm]{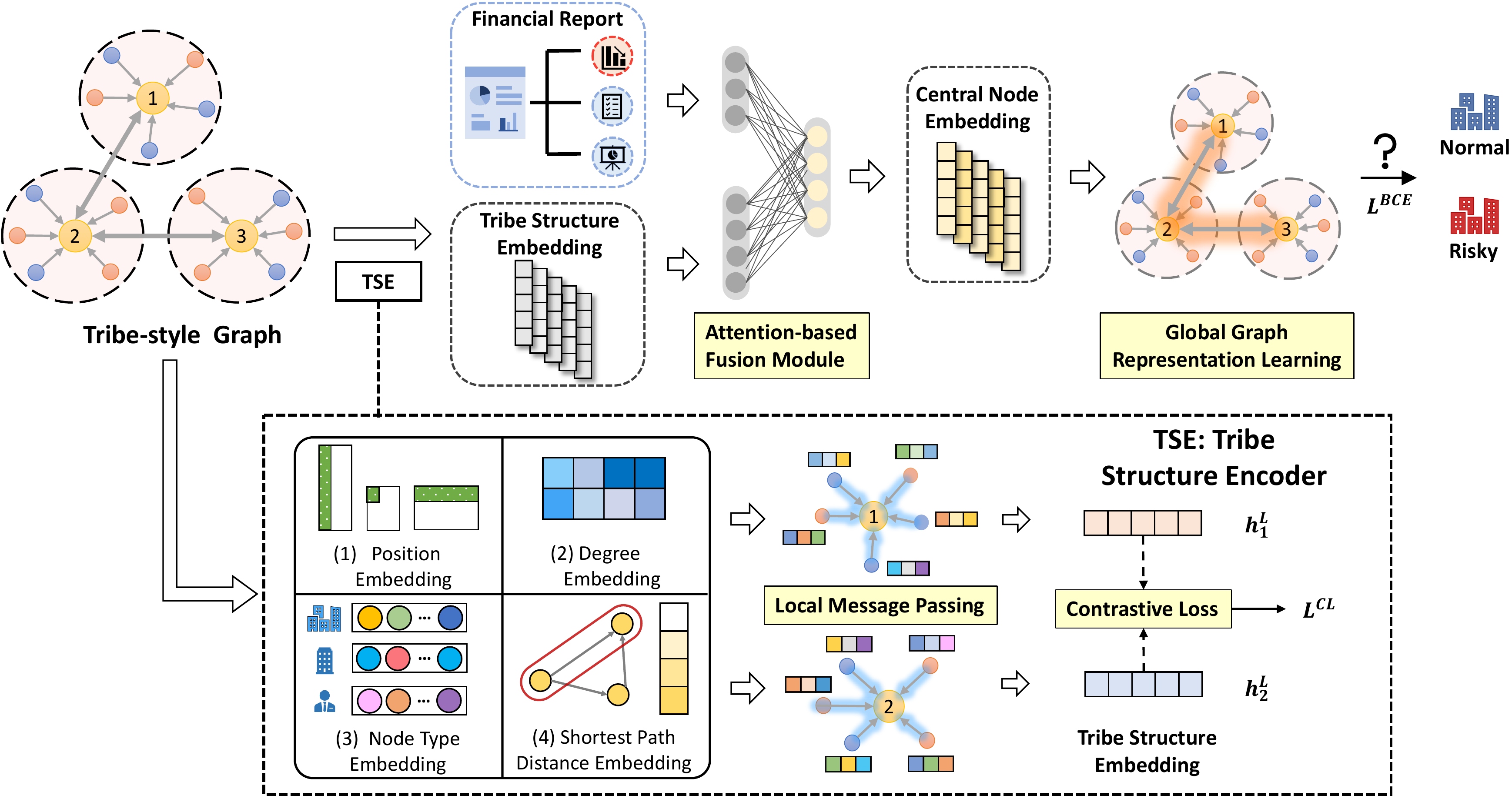}
		\caption{Overview of our proposed TH-GNN model. TH-GNN includes two main components, including the Tribe Structure Encoder (TSE) and Global Graph Represenation Learning (GGRL) module.}
		\label{fig:model_structure}
	\end{figure*}
	
	\subsection{Problem Definition}
	\label{sec:problem}
	Our company financial risk assessment problem is defined on the tribe-style graph, which consists of a set of tribes and a global news-graph connecting different tribes. For company financial risk assessment, we need to classify each company into binary classes: risky or normal. We give a formal definition of the task as follows. Let $\mathcal{G}^{\mathcal{T}}=\{\mathcal{G}^{global}, \mathcal{G}^{tribe} \}$ denote a tribe-style graph, where $\mathcal{G}^{global}=(V^G, E^G, X^G)$ is the global graph which taking all the listed companies as nodes. Considering that each listed company is the center of a tribe, we name the listed company node as the central node for shortly. $V^G$ denotes the central node set, $N^c$ is the number of central nodes, and $E^G$ is the set of edges connecting central nodes. $X^G\in \mathbb{R}^{N^c\times D}$ is the attribute matrix, and the i-th row of $X^G$ denoted by $X_i^G$ is the $D$-dimensional attribute vector of central node $v_i^G$. For each central node $v^G_i \in V^G$, there exists a tribe $g^T_i\in \mathcal{G}^{tribe}$ corresponding to $v_i^G$, where $g^T_i=(V^T_i, E^T_i)$ and $\mathcal{G}^{tribe}=\{g^T_i\ |\ i = 1\cdots N^c\}$. $V^T_i$ and $E^T_i$ are the node set and edge set of tribe $g^T_i$. Each tribe have one central node. Each central node $v_i^G$ is associated with a binary label $y_i=\{0, 1\}$, using $1$ for risky companies and $0$ for normal companies. Then the listed company's financial risks assessment problem on tribe-style graph can be described as: given a tribe-style graph $\mathcal{G}^{\mathcal{T}}$, and the goal is to classify each listed company node into binary classes: risky or normal.

	\section{Methods}
	In this section, we first explain how and why we design the tribe-style graph. Then we introduce  the proposed TH-GNN model for company financial risk assessment on tribe-style graphs in detail.

	\subsection{Tribe-style Graph Construction}
	\label{sec:graph_construct}
	\subsubsection{How to construct the tribe-style graph? }  As illustrated in Fig.~\ref{fig:data}, the tribe-style graph consists of company financial statements, investment-graphs, and financial news. Specifically, listed companies are viewed as central target nodes in the tribe-style graph, and the investment-graph of each listed company is viewed as a tribe (i.e., a super node on the tribe-style graph), where only the central listed company nodes have attributes extracted from their financial statements. The global graph connecting central companies is constructed by financial news, where two central listed companies connected if they co-existed in at least one piece of financial news. Note that we only use news that describes the risky linkages among listed companies in China. If there are multiple companies appearing in the same news, we connect all possible pairs of companies. More details about the dataset information and preprocessing are presented in Sec.~\ref{sec:dataset} and Sec.~\ref{sec:setup}.
	\subsubsection{Why we construct the tribe-style graph? }  \label{sec:benefit_tribes} We summarize the following advantages of constructing a tribe-style graph.
	\\
	(1) Based on the analysis in Sec.~\ref{section:data_analysis}, it is the structural pattern of tribes that benefits the identification of risky companies.
	\\
	(2) Considering that only the central node of a tribe has attributes, the central and non-central nodes should be treated separately. Therefore we make the tribe-style graph a hierarchical graph.
	\\
	(3) To improve model efficiency, we treat each tribe independently and obtain the representation of each tribe by graph pooling(regardless of overlap among tribes) instead of merging them into one graph, which actually truncates the original large-scale graph.
	\subsection{ Model Overview}
	We give an overview of TH-GNN in Fig.~\ref{fig:model_structure}, TH-GNN includes two main components, including the Tribe Structure Encoder (TSE) and Global Graph Representation Learning (GGRL) module. TH-GNN encodes the tribe-style graph in bottom-up order. TH-GNN first learns the structural representation for each tribe with the TSE. Then the learned structural representation of tribes and the financial statements are fused into the embedding of central node (listed company) by an attention-based fusion module. Next, the embedding is diffused over the global news-graph to learn the final representation of central nodes for financial risks assessment.

	\subsection{Tribe Structure Encoder (TSE)}
	The Tribe Structure Encoder (TSE) is used to learn the structural representation for each tribe based on contrastive learning, including a structure embedding module and a graph encoder module. Considering that nodes in the tribe have no attributes, we first initialize the node attributes according to their position in the tribe. Then we transform the structural attributes into learnable embedding with a structure embedding module for each node in the tribe. Finally, with the tribe (investment graph) and the node structure embedding, a GIN model is used to get the representation of tribes.
	
	Inspired by the importance of centrality patterns in our scenario discussed in Sec.~\ref{section:data_analysis}, we also consider the encoding of centrality when designing the TSE. For a tribe without node attributes, we first assign each node with a structure embedding as its initial attributes. Specifically, each node $v_j^T\in V_i^T$ on the investment-graph $g_i^T$ has the three properties: (1) node degree (in-degree denoted by $deg_j^+$ and out-degree denoted by $deg_j^-$ for a directed graph); (2) node type denoted by $\phi_j$ (listed company, unlisted company or human); (3) distance of shortest path (SPD) to the central node denoted by $SPD_j$. With these structure attributes, we further transform them into learnable embedding by an Embedding Layer:
	\begin{equation}
	Emb({v_j^T}) = Embedding\_Layer\left(deg_j^+, deg_j^-, \phi_j, SPD_j \right)
	\end{equation}
	Following \cite{eigenvector, qiu2020gcc}, we also use the top Laplacian eigenvector with the largest eigenvalue of each tribe as the nodes positional encoding besides the learnable structure embedding, and the $j$-th value of the top Laplacian eigenvector is exactly the eigenvector centrality \cite{bonacich1987power} of node $v_j^T \in V_i^T$. Then the final structure mebedding for each node on the tribe can be represented as:
	\begin{equation}
	\label{eq:embedding_final}
	Z(g_i^T) = \big[ Emb(v_j^T)\ ||\ u_0(g_i^T) \big]
	\end{equation}
	where $u_0(g_i^T)$ is the top Laplacian eigenvector of $g_i^T$. Next, we use the structure embedding (Eq.~\ref{eq:embedding_final}) for local message passing on tribes. In this paper, we use GIN with SUM pooling as the graph encoder for tribes. The GIN updates the node representations by:
	\begin{equation}
	h^{(l)}_{v_i^T} = MLP^{(l)}\left((1 + \epsilon^{(l)}) \cdot h^{(l-1)}_{v_i^T}  + \sum_{v_j^T\in \mathscr{N}(v_i^T) }h_{v_j^T}^{(l-1)} \right)
	\end{equation}
	where $\epsilon$ is a learnabel parameter, $h_{v_i^T}^{(l)}$ is the learned representation of node $v_i^T$ at the $l$-th layer of GIN.  Then we get the global graph representation by performing SUM Pooling for all nodes in each tribe and taking an average over all layers of the model:
	\begin{equation}
	h_{g^T_i} = \frac{1}{L+1} \sum_{l=0}^L \text{SUM}\Big( \{ h^{(L)}_{v_i^T}| v_i^T\in g_i^T \} \Big)
	\end{equation}
	Then the final tribe representation is $H_{\mathcal{G}^{tribe}} = \{h_{g^T_i} | g^T_i\in \mathcal{G}^{tribe}\}$.
	
	Considering the high cost of data labeling, we introduce contrastive learning to guide the training of TSE. 
	Specifically, we design a graph instance discrimination tasks and use InfoNCE~\cite{infoNCE} as the objective to optimize the model parameters. The contrastive task treats each tribe as a distinct class and leans to discriminate between different tribes through a self-supervised way,  and help the TSE module learn the structural dissimilarity between different tribes.
	
	Specifically, we first prepare one positive sample pair and $N-1$ negative sample pairs for each training batch with N samples and the tribes of each batch are fed into the TSE twice to obtain the query representation $H^q_{\mathcal{G}^{tribe}}$ and key representation $H^k_{\mathcal{G}^{tribe}}$ of tribes. Due to the randomness of dropout, there is a certain difference between the obtained two sets of tribe representations. Then we use the query and key representations of the same tribe as positive pairs denoted by $\{\langle q_i, k_i^+ \rangle,\ g_i^T\in\mathcal{G}^{tribe}\}$, and use the representations of different tribes to construct negative pairs denoted by $\{\langle q_i, k_j^- \rangle ,\ j\neq i\  \text{and}\ g_i^T,g_j^T\in\mathcal{G}^{tribe}\}$. Then we compute InfoNCE Loss as a regular term besides the supervised classification loss to optimize  parameters of the Tribe Structure Encoder module:
	\begin{equation}
	\label{cl_loss}
	\mathcal{L}^{CL} = -\frac{1}{N}\sum_{i=1}^N log\frac{q_i^T\cdot k_i^+}{ q_i^T\cdot k_i^+ + \sum_{j=1}^{N-1} q_i^T\cdot k_j^-}
	\end{equation}

	\subsection{Global-Graph Representation Learning (GGRL)}
	With TSE, we obtain the representations of tribes. And then for each central company, its node features come from two parts: the tribe representation and financial statements, which can be represented as $\{(h_{g^T_i}, X_i^G)\ |\ v_i^G\in V^G\}$. We further use an attention-based fusion module to integrate the tribe representations $h_{g_i^T}$ and financial statements feature $X_i^G$ into one central node embedding $h_i^{(0)}$. Finally, the fused central node embedding is used for message passing on the global news-graph to learn the final representations of central  companies.
	
	To better integrate the node features from financial statements and tribes on the global graph, we design an attention-based fusion module to fuse the two features to common space.
	We first calculate the weights for the financial statements and tribe representations:
	\begin{equation}
	\left\{
	\begin{aligned}
	e_i^g = \sigma([h_{g_i^T} \cdot W^g || X_i^G \cdot W^X] \cdot a_g^T) & , & a_g \in \mathbb{R}^{1 \times 2D} \\
	e_i^X = \sigma([h_{g_i^T} \cdot W^g || X_i^G \cdot W^X] \cdot a_x^T)  & , &  a_x \in \mathbb{R}^{1 \times 2D}
	\end{aligned}
	\right.
	\end{equation}
	where $\sigma$ is the LeakyReLU activation, $W^g$ and $W^X$ are transformation matrix to project $h_{g_i^T}$ and $X^G_i$ into common hidden dimension. Then we normalize them by the softmax function:
	\begin{equation}
	\alpha_i^g = \frac{exp(e_i^g)}{exp(e_i^g) + exp(e_i^X)},\ \alpha_i^X = \frac{exp(e_i^X)}{exp(e_i^g) + exp(e_i^X)}
	\end{equation}
	Next, we can get the fused central node embedding:
	\begin{equation}
	h^{(0)}_i = \alpha_i^g \cdot (h_{g_i^T}\cdot W^g )+ \alpha_i^X \cdot (X_i^G \cdot W^X)
	\end{equation}
	Then we use the fused central node embedding as node features and further perform message passing on the global news-graph to learn the final representations of each central node:
	\begin{equation}
	h_i^{(l)} = \sigma \left( \big(\sum_{j\in \{\mathscr{N}(v_i^G)\cup v_i^G\}} \frac{1}{d_i} \cdot h_j^{(l-1)}\big) \cdot W^l  \right)
	\end{equation}
	where $d_i$ is the in-degree of $v_i^G$ (including the self-loop). And the learned representations can be further used for the company financial risk assessment task.

	\subsection{Model Optimization Methods}
	After aggregating the information from neighbors on the graph, the obtained representation $\hat{h}_i^{(L)}$ is fed into a final fully connected neural network with a sigmoid activation function, as follows:
	\begin{equation}
	p_i = Sigmoid(h_i^{(L)} \cdot W_p + b_p)
	\end{equation}
	where $p_i$ is the probability of company node $v_i$ suffering risks in the further.
	Then we compute binary cross entropy (BCE) loss to utilize the supervised information of labels:
	\begin{equation}
	\label{bce_loss}
	\mathcal{L}^{BCE} = \frac{1}{N} \sum_{i=1}^N y_i \cdot \log \hat{y}_i + (1 - y_i) \cdot \log (1 -\hat {y}_i)
	\end{equation}
	Then, the final loss function is composed of $\mathcal{L}^{BCE}$ and $\mathcal{L}^{CL}$ (Eq.~\ref{cl_loss}):
	\begin{equation}
	\mathcal{L} = \mathcal{L}^{BCE} + \alpha \cdot \mathcal{L}^{CL}
	\end{equation}
	where $\alpha$ is a hyper-parameter to control the weight of $\mathcal{L}^{CL}$ .
	\section{Experiments}
	\label{sec:experiment}
	In this section, we compare TH-GNN with other state-of-the-art methods on a real-world dataset for company risk assessment.
	\subsection{Dataset}
	\label{sec:dataset}
	\begin{table}[h]
		\centering
		\setlength{\tabcolsep}{8.0pt}
		\caption{Information of the Hierarchical Graph}
		\label{tab:graph_info}
		\begin{tabular}{c c c}
			\toprule
			Graph & \#Nodes & \#Edges \\
			\midrule
			Whole graph $\mathcal{G}^{\mathcal{T}}$ & 879252 & 1311364 \\
			News-graph $\mathcal{G}^{global}$ & 4040 & 16330\\ 
			Investment-graphs (total)  $\mathcal{G}^{local}$ & 879252& 1295034\\
			Investment-graphs (average)  $\mathcal{G}^{local}$ & 217.6& 320.5\\
			\bottomrule
		\end{tabular}
	\end{table}
	The company dataset used in this paper comes from the real-world data of 4040 listed companies in China from 2019 to 2020, i.e., the listed company's financial statements, investment-graph, and financial news related to these companies. The financial statements and the company's investment-graph data are provided by \href{https://tianyancha.com}{TianYanCha} (an authority enterprise credit institute for company information inquiry in China). The annual financial statements reflect a listed company's industry information and its financial and business situation in a year. The investment-graph of a company describes the relationship between the central company and its shareholders, including other companies and humans. The financial news data are provided by \href{http://www.winddata.com.cn}{Wind}  ( an authority China finance database), and these news are obtained from more than 800 authority news websites in China, which have extremely wide coverage and timeliness to capture the risk information of companies. Note that the financial news used in this paper, which has already been preprocessed by \href{http://www.winddata.com.cn}{Wind}, all describes the risk linkages among companies. Then we construct the tribe-style graph  and more specific information of this graph is illustrated in Table~\ref{tab:graph_info}.
	Based on the real-world  risk events of companies happened in 2020 provided by \href{http://www.winddata.com.cn}{Wind}, the positive (risky) and negative (normal) labels can be naturally generated, and we use all companies marked as high-risk as positive samples and others as negative samples. To prevent information leakage, the part of the dataset in 2019, including financial and operating information, investment-graph and news-graph, is used as training data . There are 1698 positive samples and 2342 negative samples among all listed companies.

	\begin{table*}
		\centering
		\setlength{\tabcolsep}{10.0pt}
		\caption{Binary node classification results (\%) of models on our dataset. The model with $^*$ means using the concatenation of tribe structural embedding and financial statements data as inputs (see Sec.~\ref{sec:baseline2}). Otherwise, only financial statement data is used.}
		\label{tab:main_res}
		\begin{tabular}{ccccccc}
			\toprule
			Evaluation Metric & \multicolumn{3}{c}{binary F1-score} & \multicolumn{3}{c}{AUC score} \\
			\cmidrule(lr){2-4}\cmidrule(lr){5-7}
			Training ratio &$60\%$&$40\%$&$20\%$&$60\%$&$40\%$&$20\%$\\
			\midrule
			XGBoost &$57.4\pm1.63$ & $56.1\pm1.55$ & $55.9\pm1.05$     &$65.3\pm2.23$ &$64.3\pm2.07$ &$62.7\pm2.23$ \\ 
			DNN & $55.8\pm 2.77$ &$54.9\pm 2.56$&$53.5\pm 2.89$ &$65.1\pm 2.77$ &$65.6\pm 2.67$& $65.3\pm 2.86$\\
			GCN &$58.9\pm3.54$  &$58.3\pm1.74$&$56.8\pm2.87$ & $ 68.1\pm1.64$ &$67.2\pm1.81$&$67.8\pm1.63$ \\
			GAT & $57.6\pm3.86$ &$57.2\pm3.75$&$57.2\pm3.96$ & $67.6\pm3.11$ & $67.2\pm3.52$& $66.5\pm4.21$ \\
			GraphSAGE &$59.4\pm2.36$ &$59.5\pm2.36$&$58.5\pm2.38$  &$68.48\pm3.84$ &$67.14\pm3.25$&$65.6\pm1.78$ \\
			GCNII &$59.9\pm2.04$ & $59.7\pm1.94$&$59.1\pm1.88$  & $69.6\pm1.81$ & $69.1\pm1.72$& $68.8\pm2.01$\\
			DAGNN & $60.6\pm1.96$&$60.7\pm1.30$&\underline{$60.0\pm1.13$} & $70.6\pm1.17$ &$70.4\pm1.17$&$69.7\pm1.17$\\
			\midrule
			XGBoost$^*$ &$58.2\pm2.19$ & $56.8\pm2.05$ & $56.2\pm1.92$    &$67.2\pm2.56$ & $64.9\pm2.25$ & $64.7\pm2.33$\\ 
			DNN$^*$ & $57.2\pm2.11$ & $56.8\pm 2.32$& $55.1\pm 2.13$& $66.3\pm1.66$ &$66.4\pm 1.88$&$65.3\pm 2.15$ \\
			GCN$^*$ & $59.2\pm 1.30$&$58.6\pm 2.01$&$57.3\pm 2.01$  &$ 70.3\pm0.95$ &$68.8\pm1.31$ & $69.1\pm1.55$ \\
			GAT$^*$ & $58.4\pm2.01$& $57.4\pm2.15$& $56.8\pm2.34$  & $67.8\pm2.26$  & $67.0\pm2.34$& $66.1\pm3.12$\\
			GraphSAGE$^*$ &$59.7\pm2.08$ &$58.3\pm2.0$&$57.3\pm3.76$ &$70.6\pm2.15$ & $69.8\pm1.04$&$68.2\pm1.23$ \\
			GCNII$^*$ & $60.5\pm1.23$&$60.6\pm1.30$&\underline{$60.0\pm1.17$} & $70.9\pm1.73$ &\underline{$70.6\pm1.85$}& \underline{$70.9\pm2.09$}\\
			DAGNN$^*$ &\underline{$61.1\pm1.06$}&\underline{$60.9\pm1.12$}& {$59.9\pm3.69$}& \underline{$71.1\pm1.74$}&{$70.6\pm1.38$}&$70.2\pm0.84$ \\
			\midrule
			TH-GNN & \bm{$63.2\pm0.75$} &\bm{$62.8\pm0.95$}&\bm{$62.2\pm1.13$} &  \bm{$73.5\pm0.54$} &\bm{$72.8\pm0.54$}&\bm{$72.5\pm0.54$} \\
			\bottomrule
		\end{tabular}
	\end{table*}
	
	\subsection{Experimental Setup}
    \label{sec:setup}	
	We conduct experiments on the real-world dataset with different configurations. We design experiments with different training ratios (percent of nodes in the training set) ranging from 20\% to 40\%. And for each training ratio,  we use three different random partitions of the dataset and ten random seeds for the model parameter initialization, a total of 30 trials for each model.  For all attributes of the dataset used in this paper, we preprocess the categorical or discrete attributes into one-hot vectors, and then we use binning methods to divide continuous numerical attributes into 50 bins and use the index of bin as their feature. For fairness, we perform a hyper-parameter search for all models, and the size of searching space for each model is the same.  The hidden dimension of all models are searched in \{32, 64, 128\} and we choose the number of training epoch from \{100, 200, 300\}. We use the Adam optimizer for all experiments and the learning rate is searched in \{1e-2, 1e-3, 1e-4\}, weight decay is searched in \{1e-4, 1e-3, 5e-3\}, and $\alpha$ (the coefficient of $\mathcal{L}^{CL}$) is searched in \{0.01, 0.05, 0.1, 0.5, 1.0\} for all experiments. The number of layers for GNN models, including TH-GNN and other baseline GNN models except for GCNII and DAGNN, are set to be two layers in this paper. The number of layers for GCNII and DAGNN, which are designed with deeper depth, are set to 64 and 20 respectively according to their papers \cite{GCN2,DAGNN}. All models used in this paper were trained on Nvidia Tesla V100 (32G) GPU.
	
	\subsection{Compared Methods}
	\subsubsection{Baseline methods}
	\label{sec:baseline}
	We compare our model with two classical machine learning models (XGBoost~\cite{chen2016xgboost}, DNN), three baseline GNN models (GCN~\cite{GCN}, GAT~\cite{GAT}, GraphSAGE~\cite{GraphSAGE}), and two state-of-the-art GNN models (GCNII~\cite{GCN2}, DAGNN~\cite{DAGNN}) to demonstrate the superiority of our TH-GNN model. Furthermore, six variants of TH-GNN are designed for ablation studies: $\text{TH-GNN}_{\backslash\text{Attribute}}$  indicates  removing node attributes (financial statements) from TH-GNN and only using the graph structure information; 
	$\text{TH-GNN}_{\backslash\text{TSE}}$ indicates removing Tribe Structure Encoder from TH-GNN, without leveraging tribes; 
	$\text{TH-GNN}_{\backslash\text{GGRL}}$ indicates removing the GGRL module from TH-GNN, without leveraging the news-graph;
	$\text{TH-GNN}_{\backslash\text{Fusion}}$ indicates removing attention-based fusion module from TH-GNN;
	$\text{TH-GNN}_{\backslash\text{Emb}}$ indicates removing the structure embedding used in TSE;
	$\text{TH-GNN}_{\backslash\text{CL}}$ indicates removing the contrastive loss term used to optimize the Tribe Structure Encoder from TH-GNN. 	In our experiments, we select two widely used metrics as performance measurement, i.e., AUC (the area enclosed by the coordinate axis under the ROC curve) and F1-score (the harmonic average of the precision and recall) on the test set.

	\subsubsection{Two implementations for base GNN models}
	\label{sec:baseline2}
	Note that GNN models except TH-GNN cannot directly encode the  tribe-style graph with hierarchical structures. For fair comparisons,  we design two implementations for each baseline GNN model.
	
	(1) \textbf{A basic implementation} is to directly train the vanilla baseline GNNs on the {news-graph}, because nodes except the central node (listed company) in a tribe do not have attributes (financial statements), which are necessary for vanilla GNN models.
	
	(2) \textbf{A two-stage implementation} is to learn structure representation of tribes (investment-graphs) first and then train baseline GNNs on the news-graph. In this paper, we use GCC~\cite{qiu2020gcc} , a graph contrastive learning based method, to learn structure representation for tribes, which are concatenated with the financial statement feature as attributes of nodes on the news-graph.

	
	\subsection{Main Results}
	The main results of different models are presented in Table~\ref{tab:main_res} and the major findings are summarized as follows:
	
	(1) We observe that our TH-GNN model significantly outperforms other competing methods. Its binary F1-score, with the reported value of 63.2, is at least 5\% higher than the tree-based model and traditional DNN model, and AUC gets 6.3\%  higher at the same time. Furthermore, TH-GNN is more advanced than the state-of-the-art GNN-based methods, i.e., GCNII and DAGNN, with about 2.1\% increased F1-score and 2.4\% increased AUC. Besides, the lower standard deviations of the results of TH-GNN indicate that our proposed model is more robust with different dataset splits. 
	
	(2) The results on the upper and lower sides of the middle line in Tab.~\ref{tab:main_res} show the comparison of the two implementations of baselilne models (see Sec.~\ref{sec:baseline2}). Generally, the performance of base models with tribe structure representations as additional input (the model with  $^*$)  are improved with varying degrees, which demonstrate the effectiveness of tribe structure encoding. Besides, TH-GNN outperforms the state-of-the-art GNN methods with tribe structure encoding as additional input. These improvements are mainly brought by the TSE module of TH-GNN that considers encoding of graph centrality and the end-to-end training strategy under the guidance of the supervised loss and the contrastive  loss  jointly.
	
	\begin{table}[h]
		\centering
		\setlength{\tabcolsep}{10.0pt}
		\caption{Binary node classification results (\%) of TH-GNN and its variants (see Sec.~\ref{sec:baseline}) for ablation study.}
		\label{tab:ablation}
		\begin{tabular}{ccc}
			\toprule
			Graph & binary F1-score & AUC \\
			\midrule
			$\text{TH-GNN}_{\backslash\text{Attribute}}$ & $56.7_{\backslash-6.5}$ & $62.4_{\backslash-10.5}$ \\
			$\text{TH-GNN}_{\backslash\text{GGRL}}$ & $58.2_{\backslash-5.0}$ & $67.4_{\backslash-6.1}$ \\ 
			$\text{TH-GNN}_{\backslash\text{TSE}}$ & $60.2_{\backslash-3.0}$ & $69.5_{\backslash-4.0}$ \\ 
			$\text{TH-GNN}_{\backslash\text{CL}}$ & $62.0_{\backslash-1.2}$ & $71.9_{\backslash-1.6}$ \\
			$\text{TH-GNN}_{\backslash\text{Emb}}$ & $62.6_{\backslash-0.6}$ & $72.1_{\backslash-1.4}$ \\ 
			$\text{TH-GNN}_{\backslash\text{Fusion}}$ & $62.5_{\backslash-0.7}$ & $71.6_{\backslash-1.9}$ \\ 
			
			\midrule
			$\text{TH-GNN}$ & \bm{$63.2_{\backslash{0.0}}$} &  \bm{$73.5_{\backslash{0.0}}$} \\
			\bottomrule
		\end{tabular}
	\end{table}
	
	\subsection{Ablation Study}
	Then, we perform ablation studies to demonstrate the effectiveness of every component in our model. As shown in Table~\ref{tab:ablation}, the binary F-1 and AUC of all variants deteriorate to some extent.

	\subsubsection{The effects of tribe-style graph}
	First, we demonstrate the effects of the tribe-style graph by removing information of certain type (i.e.,  the financial statements,  global news-graph or local tribes) respectively. $\text{TH-GNN}_{\backslash\text{Attribute}}$, as a variant without financial statements as node attributes, obtains the worst performance among all the variants with 6.5\% decreased in F1-score and 10.5\% decreased in AUC. Moreover, the reasonable score shows that even without the guidance of financial statements, only the graph structure information can still guarantee a certain classification accuracy, which demonstrates the effectiveness of our proposed tribe-style structure. Moreover, the performance degradation of $\text{TH-GNN}_{\backslash\text{GGRL}}$ and $\text{TH-GNN}_{\backslash\text{TSE}}$ indicates that the global news-graph and local tribes (investment-graphs) both benefit the downstream task.
	
	\subsubsection{The effects of different model components}
	To further verify the importance of different components in our model, another three variants of TH-GNN are designed for ablation study.
	The performance degradation of $\text{TH-GNN}_{\backslash\text{CL}}$, which removes the contrastive loss term, shows that the contrastive loss $L^{CL}$ plays an essential role in learning the discriminative structural information of tribes.
	Besides, the whole TH-GNN model yields a good performance boost in comparison to  $\text{TH-GNN}_{\backslash\text{Emb}}$, a variant without the structure embedding in TSE and using randomly initialized attributes for nodes on tribes, which indicates that the proposed structure embedding is beneficial for modeling graphs without node attributes.
	From the results of $\text{TH-GNN}_{\backslash\text{Fusion}}$, we find that the fusion module, which integrates the information from financial statements and tribes comprehensively, brings significant improvements. Overall, the whole TH-GNN achieves the best result compared with all variants.

	\subsection{Visualization of Learned Representations}
	
	
	
	\begin{figure}[h]
		\begin{minipage}[t]{0.32\linewidth}
			\centering
			\subfloat[GCN]{\includegraphics[width=\linewidth, height=2cm]{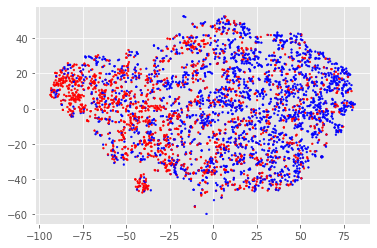}}
		\end{minipage}
		\begin{minipage}[t]{0.32\linewidth}
			\centering
			\subfloat[GAT]{\includegraphics[width=\linewidth, height=2cm]{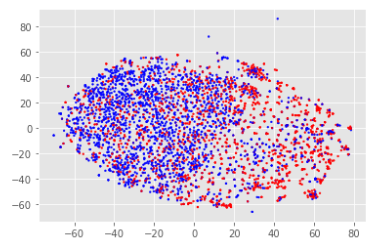}}
		\end{minipage}
		\begin{minipage}[t]{0.32\linewidth}
			\centering
			\subfloat[GraphSAGE]{\includegraphics[width=\linewidth, height=2cm]{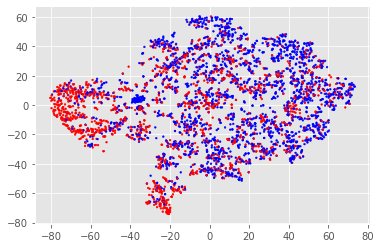}}
		\end{minipage}
		\begin{minipage}[t]{0.32\linewidth}
			\centering
			\subfloat[GCNII]{\includegraphics[width=\linewidth, height=2cm]{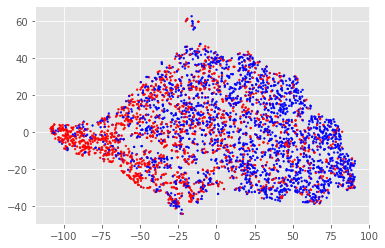}}
		\end{minipage}
		\begin{minipage}[t]{0.32\linewidth}
			\centering
			\subfloat[DAGNN]{\includegraphics[width=\linewidth, height=2cm]{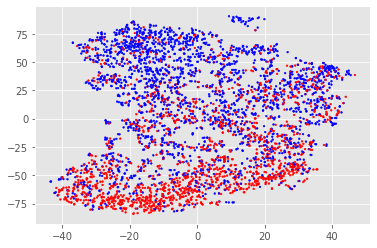}}
		\end{minipage}
		\begin{minipage}[t]{0.32\linewidth}
			\centering
			\subfloat[TH-GNN]{\includegraphics[width=\linewidth, height=2cm]{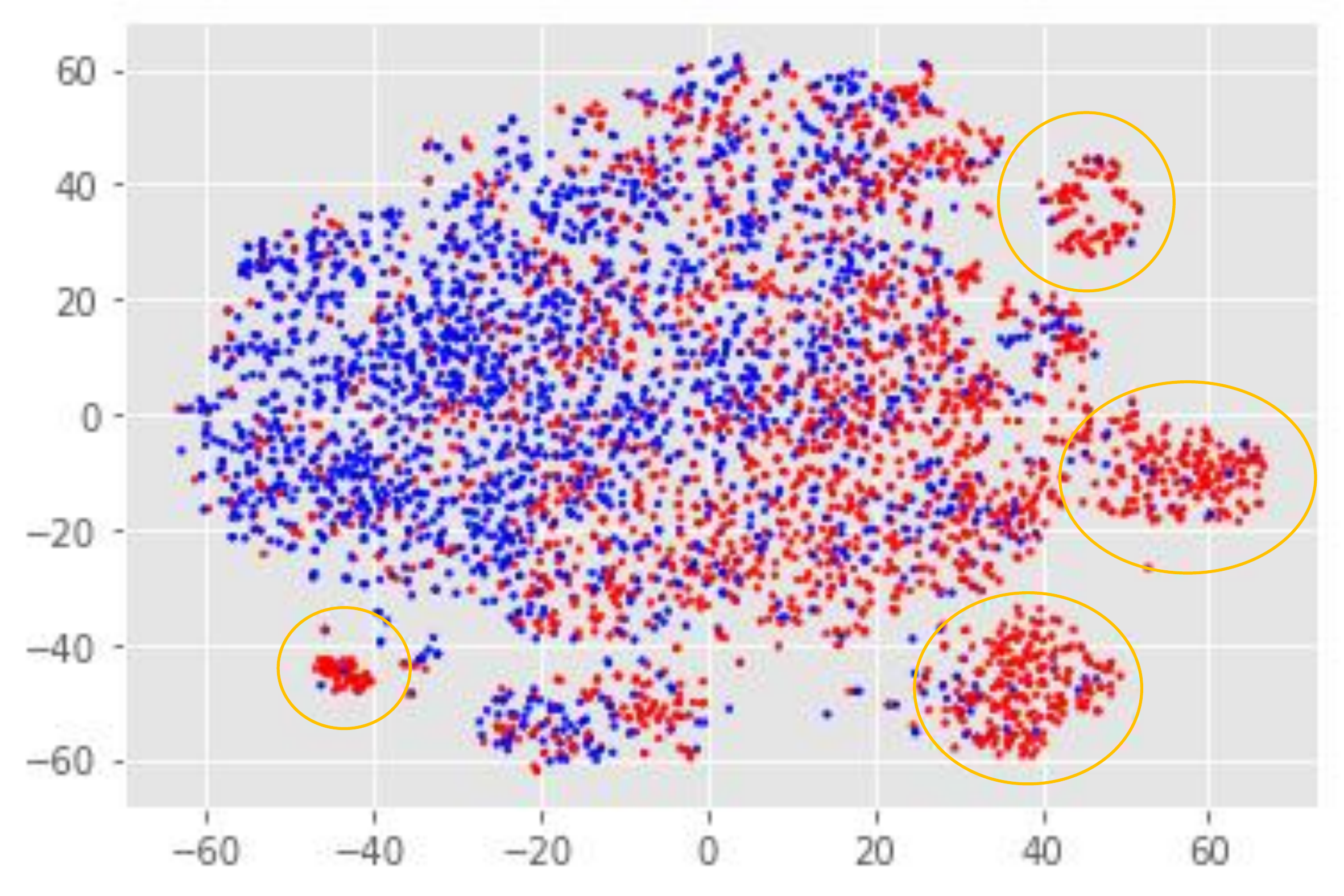}}
		\end{minipage}
		
		\caption{T-SNE visualization of the representations learned by models, where red dots represent samples of risky companies and blue dots represent samples of normal companies.}
		\label{fig:tsne}
	\end{figure}
	
	We analyse the model's interpretability by visualizing the learned representations. We first extract the hidden outputs of the penultimate layer for different GNN models. Next, we reduce the hidden vectors to 2-dimensional vectors with T-SNE algorithm and plot their the scatter plot . 
	As shown in Fig.~\ref{fig:tsne}, the representations learned by GCN, GAT and GraphSAGE have weak discrimination between risky and normal nodes. Despite the representations learned by GCNII and DAGNN have improved node classification ability, our TH-GNN has the best discriminative power and provides interpretability that other GNN models do not have. Fig.~\ref{fig:tsne} (f) illustrates the visualization of the representations learned by TH-GNN which have evident outlier clusters, and most of points in these clusters are risky companies. This unique phenomenon indicates that some risky companies have similar local investment-patterns (i.e., the example presented in Fig.~\ref{fig:subgraph_examples} (a)), which is easily affected by their risky neighbors. And this observation further demonstrate the indispensable effects of the Tribe Structure Encoder (TSE).

	
	\section{Related Work}
	In this paper, we mainly focus on using graph learning methods to solve the company financial risk assessment problem.
	
	\textbf{Graph Representation Learning.}\quad Graph Neural Networks (GNNs) \cite{GCN,gwnn,xu2020graph,GAT,GraphSAGE,xu2020survey} have shown excellent  performance on graph representation learning. Graph Convolution Network (GCN)~\cite{GCN} was first proposed with aggregation-based graph convolution operation. After that, its improved models have been proposed. GAT~\cite{GAT} introduced an attention mechanism to distinguish the importance of neighbors. GIN~\cite{GIN} investigate the effect of different aggregation functions for graph representation ability. GraphSAGE~\cite{GraphSAGE} proposed to use graph sampling for inductive learning on graphs. Recently, some studies \cite{GCN2,DAGNN} also focus on deeper GCN with larger receptive filed. GCNII~\cite{GCN2} is an extension of vanilla GCN with initial residual and identity mapping.  DAGNN~\cite{DAGNN} is a deeper GNN model that decouples the transformation and aggregation of message passing. However, most of them \cite{xu2020graph, xu2020label,fu2021neuron, mao2021neuron,mao2021source} are designed without taking the unique domain knowledge into consideration and are not aimed at tackling the company financial risk assessment on financial networks.
	
	\textbf{Company Financial Risk Assessment.} \quad
	Financial risk has been a major concern for financial companies and governments, and extensive works have been studied to assess company financial risks. With the rapid development of machine learning methods in recent years, researchers have developed machine learning-based company financial risk prediction models using financial statements or privately-owned data provided by financial institutions \cite{mai2019deep, chen2020ensemble, erdogan2013prediction, zhang2021form}.  
	After the global financial crisis in 2008, the crisis brought by the collapse of Lehman Brothers spread rapidly and widely to related companies, and graph theory began to attract the attention of researchers \cite{meng2017netrating, niu2018visual,allen2009networks, bougheas2015complex, trkman2009supply, rezaei2019neural, Sun2011trading, Sun2014predict, xu2021towards,liu2021pick}. 
	Recently, some studies attempt to construct different financial graphs and design domain-specific GNNs for company financial risk assessment \cite{feng2020every, cheng2019risk, yang2020financial, zheng2021heterogeneous}. For example, \cite{cheng2019risk} proposed a High-order Graph Attention Networks to assess risks for companies on guarantee loans networks considering higher-order neighbors. \citet{yang2020financial} proposed a dynamic GNN for supply chain mining. \citet{zheng2021heterogeneous} designed a heterogeneous graph attention network to predict the bankruptcy risks of small and medium-sized companies. However, all these works only consider one type of financial graph, and the data is specific and private. Models trained on such domain-specific data may have bias and have poor generalization ability. 
	\section{Conclusion and further work}
	In this paper, we investigate the listed company's financial data in real words and find that it is far from sufficient to assess the risk of listed companies solely through financial statements. To provide more comprehensive representations of companies, we design a tribe-style network and propose TH-GNN for company financial risk assessment task. Experiments on a real-world large-scale dataset show that our proposed model is effective in company financial risk assessment task and can provide brilliant interpretability of results. Furthermore, much future work is focused on the scalability of the method to investigate problems such as credit evaluation, risk transmission and so on. In addition, TH-GNN provides valuable tools to analyse the information of companies and their risks. As future work, we will try more complicated GNN backbones rather than simply GIN and GCN,  and combine real-time information with tribe-style graph to further improve the dynamic risk identification.
	\begin{acks}
		This work was supported by the National Natural Science Foundation of China (Grant No.61902380, No.61802370, No.U21B2046 and No.U1911401) and the Beijing Nova Program (No. Z201100006820061).
	\end{acks}
	\bibliographystyle{ACM-Reference-Format}
	\bibliography{sample-base}
	
	\appendix
	
\end{document}